\documentclass[letterpaper, 10 pt, conference]{IEEEtran}
\IEEEoverridecommandlockouts
\usepackage[colorlinks=true,allcolors=black,breaklinks]{hyperref}
\usepackage[switch]{lineno}
\usepackage{algorithm}
\usepackage{algorithmic}

\usepackage{geometry}
\usepackage{amsmath}
\usepackage{amssymb}
\usepackage[noabbrev,capitalize]{cleveref}
\usepackage{thmtools}
\usepackage{complexity}
\usepackage{cite}
\usepackage[usenames, dvipsnames]{color}
\usepackage{graphicx}
\usepackage{siunitx}
\usepackage{lmodern}
\usepackage{subcaption}

\DeclareCaptionLabelFormat{opening}{#2)}
\usepackage[font=small]{caption}
\usepackage{balance}
\usepackage{xspace}
\usepackage[export]{adjustbox}
\usepackage[low-sup]{subdepth}
\graphicspath{{figures/}}

\usepackage[font=footnotesize]{caption}
\usepackage{float}
\usepackage{textcomp}

\newcommand{\BILLE}{\makebox{\textsc{Bill-E}}\xspace}
\newcommand{\SOLLE}{\makebox{\textsc{Soll-E}}\xspace}
\newcommand{\ARMADAS}{\makebox{\textsc{Armadas}}\xspace}
\newcommand{\CHHC}{\makebox{CH\textsubscript{2}C}\xspace}
\newcommand{\GLC}{GLC\xspace}
\newcommand{\MWPMexp}{\textsc{MWPMexpand}\xspace}
\newcommand{\boxy}{\textsc{Boxy}\xspace}
\newcommand{\snakey}{\textsc{Snakey}\xspace}
\newcommand{\overlapping}{\textsc{Overlapping}\xspace}

\title{\LARGE\bf  Moving Matter: Using a Single, Simple Robot to Reconfigure a Connected Set of Building Blocks\vspace{-0.5em}}
\author{Javier Garcia\textsuperscript{1},
    Jonas Friemel\textsuperscript{2},
    Ramin Kosfeld\textsuperscript{3},
    Michael Yannuzzi\textsuperscript{1},
    Peter Kramer\textsuperscript{3},\\
    Christian Rieck\textsuperscript{4},
    Christian Scheffer\textsuperscript{2},
    Arne Schmidt\textsuperscript{3},
    Harm Kube\textsuperscript{5},
    Dan Biediger\textsuperscript{1},\\
    Sándor P.\ Fekete\textsuperscript{3}, and
    Aaron T.\ Becker\textsuperscript{1}
\thanks{\textsuperscript{1}Partially supported by NSF \href{https://www.nsf.gov/awardsearch/showAward?AWD_ID=2130793}{IIS-2130793},  ARL W911NF-23-2-0014, and DAF AFX23D-TCSO1.
Electrical Engineering, University of Houston, Texas, USA, \{atbecker, mcyannuz, jgarciag, debiedig\}@cougarnet.uh.edu}
\thanks{\textsuperscript{2}Partially supported by DFG grant SCHE~1931/4-1.
Electrical Engineering and Computer Science, Bochum University of Applied Sciences, Germany,  \{jonas.friemel, christian.scheffer\}@hs-bochum.de}
\thanks{\textsuperscript{3}Partially supported by DFG grant FE~407/21-1.
Computer Science, TU Braunschweig, Germany, \{kosfeld, kramer, aschmidt\}@ibr.cs.tu-bs.de, s.fekete@tu-bs.de}
\thanks{\textsuperscript{4}Partially supported by DFG grant~522790373.
Discrete Mathematics, University of Kassel, Germany, christian.rieck@mathematik.uni-kassel.de}
\thanks{\textsuperscript{5}Computer Science, TU Berlin, Germany, h.kube@campus.tu-berlin.de}
}

\begin{document}
    \maketitle
    \begin{abstract}
        We implement and evaluate different methods for the reconfiguration of a connected arrangement of tiles into a desired target shape, using a single active robot that can move along the tile structure.
        This robot can pick up, carry, or drop off one tile at a time, but it must maintain a single connected configuration at all times.

        Becker et al.~\cite{becker2025moving} recently proposed an algorithm that uses histograms as canonical intermediate configurations, guaranteeing performance within a constant factor of the optimal solution if the start and target configuration are well-separated.
        We implement and evaluate this algorithm, both in a simulated and practical setting, using an inchworm type robot to compare it with two existing heuristic algorithms.
    \end{abstract}

    \section{Introduction}

    \begin{figure}[!t]
        \centering
        \includegraphics[width=\columnwidth]{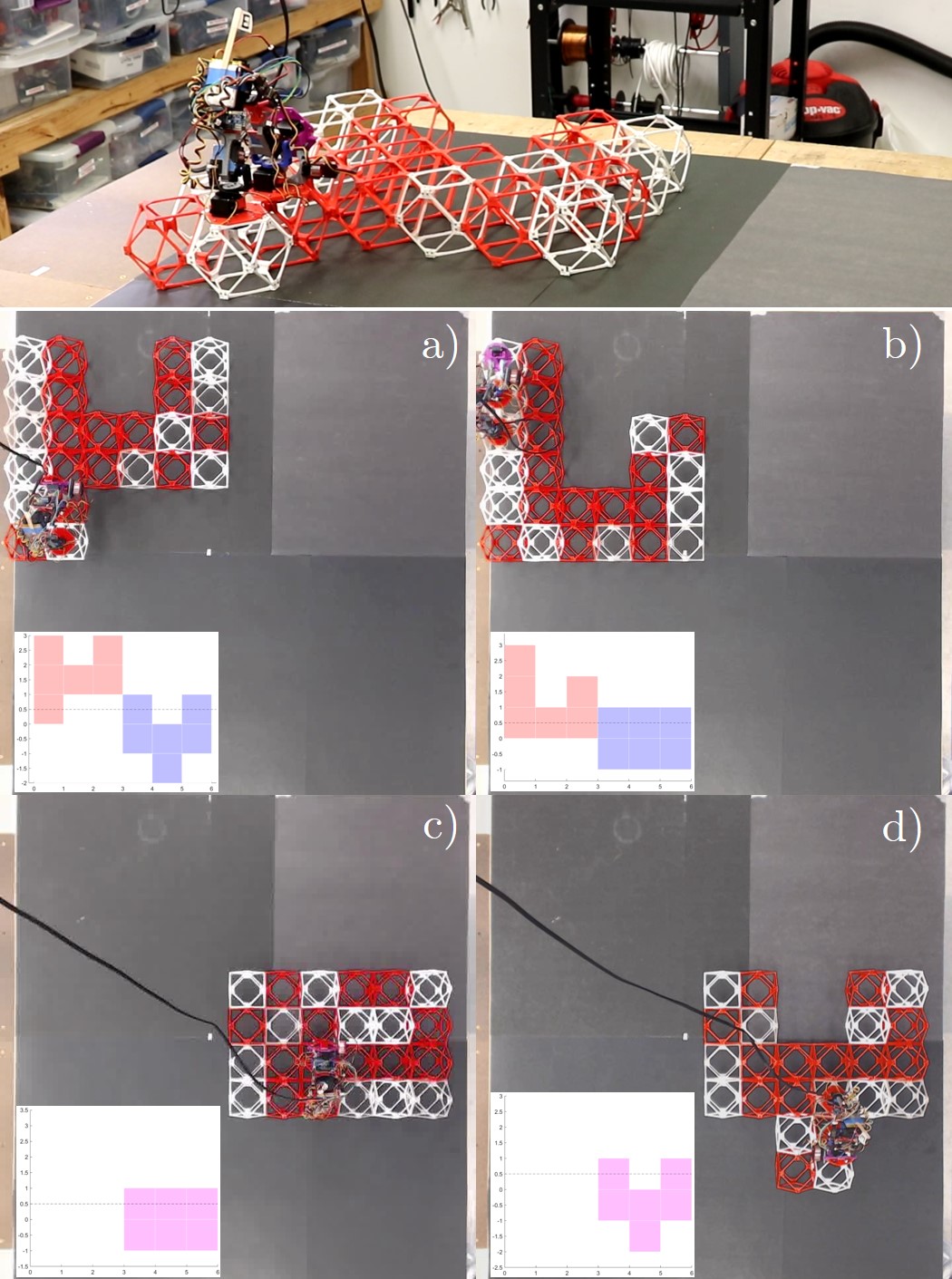}
        \caption{\label{fig:conceptImg} Demonstration of a reconfiguration using the proposed algorithm.
a) The starting (red) and target (blue) configurations.
b) The configurations are transformed into histograms.
c) The starting histogram is moved to match the target histogram (purple because they overlap).
d) The histogram process is reversed to arrive at the desired target configuration.
Connectivity of the structure is maintained at all times.
See the full reconfiguration at \url{https://youtu.be/0jE2TiQfYJs}.
        }
    \end{figure}

    Since antiquity, humans have repurposed building materials from abandoned structures, infamously using the Colosseum in Rome as a quarry for new projects~\cite{hopkins2012colosseum}.
Reusing the nearby, regular stones was easier than mining and shaping new ones.
    In programmable matter, reconfiguring modular structures remains a challenge of ongoing research.
    While many works have explored systems of ``active'' intelligent units capable of adjusting their own shape~\cite{doi:10.1177/02783649241283847}, for large-scale applications, it may be more efficient to use passive building materials that are manipulated by a small set of active robots.
    There has been significant progress in both macroscopic and microscopic domains, e.g., for building large manufacturing structures in space~\cite{jenett2017design} or underwater~\cite{LensgrafConstruction2023}, as well as nano-scale assembly using DNA~\cite{song2017reconfiguration}.

    In the \BILLE model~\cite{jenett2017bille,jenett2019material}, an inchworm \mbox{6-DoF} robot can walk on a connected structure of voxels (or tiles) that are aligned on a lattice; see~\cref{fig:conceptImg} for an illustration.
    Similarly to the \SOLLE robots developed by NASA for the \ARMADAS platform~\cite{costa2019algorithmic,ParkCFGOOSSTTC23,ultralight24}, \BILLE robots are not only able to walk on top of the configuration, but can also lift and place tiles, enabling autonomous assembly without relying on external workspaces as, e.g., in 3D printing setups.
    On these platforms, a natural task is to reconfigure between tile arrangements.
    With aspects like power consumption and build time in mind, we aim for fast reconfiguration schedules.
    As disconnected tiles may drift apart in environments like oceans or space, tiles should be connected at all times.

    Focusing on the \BILLE platform, several single-robot reconfiguration algorithms have been developed:
    Garcia et al.~\cite{single-bille-reconfig-IROS} introduced and evaluated an algorithm that iteratively grows the largest connected component (\GLC) of tiles at the target position, as well as two heuristics based on pair matchings (\MWPMexp),
    and rapidly exploring random trees~(RRT).
    In a follow-up publication, they studied the respective multi-robot variant~\cite{cooperative-bille-reconfig-ICRA}.
    Most recently, \mbox{Becker et al.~\cite{becker2025moving}} proposed a constant factor approximation algorithm for reconfiguring between configurations of disjoint bounding boxes in a model similar to the \BILLE.

    \subsection{Our Contribution}\label{subsec:our-results}

    In this paper, we investigate the problem of finding minimum-cost reconfiguration schedules for a single active
    robot operating on a (potentially large) number of tiles.
    In particular, we present a practical evaluation of the aforementioned approximation algorithm~\cite{becker2025moving}, comparing its performance with both the \GLC and \MWPMexp algorithms from earlier work~\cite{single-bille-reconfig-IROS} in terms of total costs for different sets of maps, as well as other metrics such as the number of pickup/dropoff pairs.
The algorithm is also physically implemented to demonstrate its feasibility.

    \subsection{Related Work}\label{sec:RelatedWork}
    Garcia et al.~\cite{cooperative-bille-reconfig-ICRA} show that deciding the length of optimal schedules to reconfigure polyominoes with \BILLE bots is in general \mbox{\NP-hard}.
    For the slightly broader model that is also used in this work, Becker et al.~\cite{becker2025moving} prove that deciding the minimal schedule length for reconfiguration remains \NP-hard, even if movement cost is higher while carrying tiles.

    To address this, Garcia et al.~\cite{single-bille-reconfig-IROS} present heuristic approaches exploiting RRT combined with a time-dependent variant of the A$^{*}$ search algorithm.
    Furthermore, they use target swapping heuristics to reduce the overall traveled distance in the case of multiple \BILLE bots~\cite{cooperative-bille-reconfig-ICRA}.
    Becker et al.~\cite{becker2025moving} propose a deterministic algorithm that reconfigures between two connected configurations with a performance that is within a constant of an optimal schedule.
    As this paper is based on these approaches, we discuss them in more detail in~\cref{sec:approx,sec:simulation}.

    A different context arises from programmable matter~\cite{friemel2025reconfiguration,gmyr2020forming,hinnenthal2024efficient}.
    Here, even finite automata are capable of building bounding boxes from tiles
    around polyominoes, as well as scaling and rotating them while maintaining
    connectivity at all times~\cite{fekete2022connected,NiesReconfig}.

    When considering active matter, i.e., the arrangement is composed of self-moving objects (or agents), several models exist~\cite{almethen2020pushing,almethen2022efficient,connor2025transformation,michail2019transformation}.
    For~example, in the \emph{sliding square model}~\cite{FitchBR03,FitchBR05} agents are allowed to make two different moves, namely, sliding along other agents or performing a convex transition, i.e., moving around a convex corner.
    Akitaya~et~al.~\cite{AkitayaDKKPSSUW22} show that universal sequential reconfiguration in two dimensions is possible, even while maintaining connectivity of all intermediate configurations, but minimizing the schedule size is \NP-complete.
    Most recently, the authors of~\cite{parallel-sliding-squares} show that it is \mbox{\NP-complete} to decide whether there exists a schedule of length~$1$ in the parallel sliding square model and provide a worst-case optimal reconfiguration algorithm.
    The MIT--NASA Space Robots Team present a reconfiguration algorithm for three-dimensional sliding cubes with movement constraints which ensure that the algorithm can be executed by \SOLLE robots~\cite{TeamBCDDGHK24}.

    \subsection{Preliminaries and Definitions}\label{sec:prelims}

    In this work, we consider a simplified material-robot model as used by Becker et al.~\cite{becker2025moving}.
    We work on the two-dimensional infinite integer grid~$\mathbb{Z}^2$ where some vertices are occupied by indistinguishable square tiles.
    We say that two grid positions are \emph{adjacent} if they are contained in each other's 4-neighborhood.
    A set of tiles is \emph{connected} if the adjacency graph of all tiles is connected.
    We call a connected set of tiles a \emph{configuration} or \emph{polyomino} and refer by $C(n)$ to the set of all configurations consisting of $n \in \mathbb{N}$ tiles.

    We assume that a single robot occupies exactly one tile of a configuration.
    The robot supports three different \emph{operations}:
    it can \emph{move} from one tile to an adjacent one,
    \emph{pick up} an adjacent tile (if this does not break connectivity)
    or \emph{drop off} a tile at an adjacent unoccupied position.
    In our model, the robot is capable of carrying at most one tile at a time, so if we neglect the move operations, a start configuration $C_s\in C(n)$ is reconfigured into a target configuration $C_t\in C(n)$ by an alternating sequence of pickup and dropoff operations.
    Such a sequence is called a \emph{schedule}.
    The \emph{makespan} of a schedule~$S$ is the number of operations required to realize~$S$, where pickup, dropoff, and movement are weighted equally.

    \section{Approximation Algorithm}
    \label{sec:approx}

    \begin{figure}[t]
        \centering
        \includegraphics[width=\columnwidth, trim={0 0 0 0},clip]{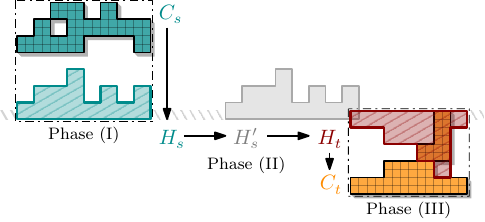}
        \caption{
            \label{fig:alg-overview}
            The approximation algorithm by Becker et al.~\cite{becker2025moving} performs a reconfiguration from $C_s$ to $C_t$ in three phases, using two histograms $H_s$ and $H_t$ as intermediate configurations.
            Phases (I) and (III) transform between a histogram and start/target configuration, Phase (II) transforms between two histograms facing opposing directions.
        }
    \end{figure}

    Complementarily to establishing \NP-hardness, Becker et al.~\cite{becker2025moving} present an algorithm for a single robot to reconfigure connected tile configurations with disjoint bounding boxes.
    For most of the work, tile configurations are \emph{2-scaled}, i.e., they consist of $2\times 2$ tiles rooted in even grid coordinates.
    While this is the first algorithm in this setting that generates schedules with a makespan within a constant factor of an optimal solution, its practical performance in comparison to existing reconfiguration algorithms has not yet been evaluated.

    We give a high-level overview of the algorithmic ideas and refer to~\cite{becker2025moving} for details.
    The algorithm is divided into three phases, and makes use of a class of intermediate shapes called \emph{histograms}.
    A~histogram consists of a unit-height \emph{base} strip and multiple unit-width \emph{columns} orthogonally attached to the base.
    In the first phase, the initial configuration~$C_s$ is transformed into a histogram~$H_s$.
    In the second phase, $H_s$ is transformed into a histogram~$H_t$ contained in the bounding box of the target configuration~$C_t$.
    Finally, in the third phase, $H_t$ is transformed into $C_t$ using the same technique as for the first phase, but executing the schedule in reverse; see \cref{fig:alg-overview}.

    Note that the algorithm mainly relies on two subroutines; one for transforming configurations into histograms (Phase (I), Phase (III) time-reversed), and another for reconfiguring between two histograms (Phase (II)).
    We describe both techniques:

    \subsection{Creating Histograms, Phases (I) and (III)}
    \label{subsec:creating-histograms-phases-(i)-and-(iii)}
    Without loss of generality, we assume that the bounding boxes of the start and target configuration are separated by a horizontal bisector, i.e., tiles can be split into start and target configurations depending on a threshold value for their $y$-coordinate, see \cref{fig:bounding-box-overlap-horizontal-bisector}.
    If the configurations are instead separated by a vertical bisector (\cref{fig:bounding-box-overlap-vertical-bisector}), we rotate the map by 90 degrees.
    To create histograms, we first select a vertical position for the histogram's base, followed by identifying certain parts of the configuration, referred to as \emph{free components}, that can be moved without violating the connectivity constraint; the algorithm then iteratively shifts these parts downwards to obtain the desired histogram.

    The downward movement is performed by a subroutine that visits every column in a free component, moving the topmost tile of each column down.

    \subsection{Reconfiguring Between Histograms, Phase~(II)}
    \label{subsec:reconfiguring-between-histograms-phase(ii)}
    To reconfigure between two histograms that face opposing directions, we assume that both histograms share at least one tile.
    Otherwise, the subroutine for moving free components can be applied first to shift the histogram $H_s$ horizontally (e.g., $H_s'$ in \cref{fig:alg-overview}).
    The reconfiguration proceeds by iteratively moving the topmost, leftmost tile in $H_s - H_t$ to the topmost, leftmost target position in $H_t - H_s$.

    \section{Simulations}
    \label{sec:simulation}

    \subsection{Simulation Methods}
    \label{subsec:simulation-methods}

    \begin{figure}[t]
        \centering
        \includegraphics[width=.8\columnwidth, trim={0 0 0 0},clip]{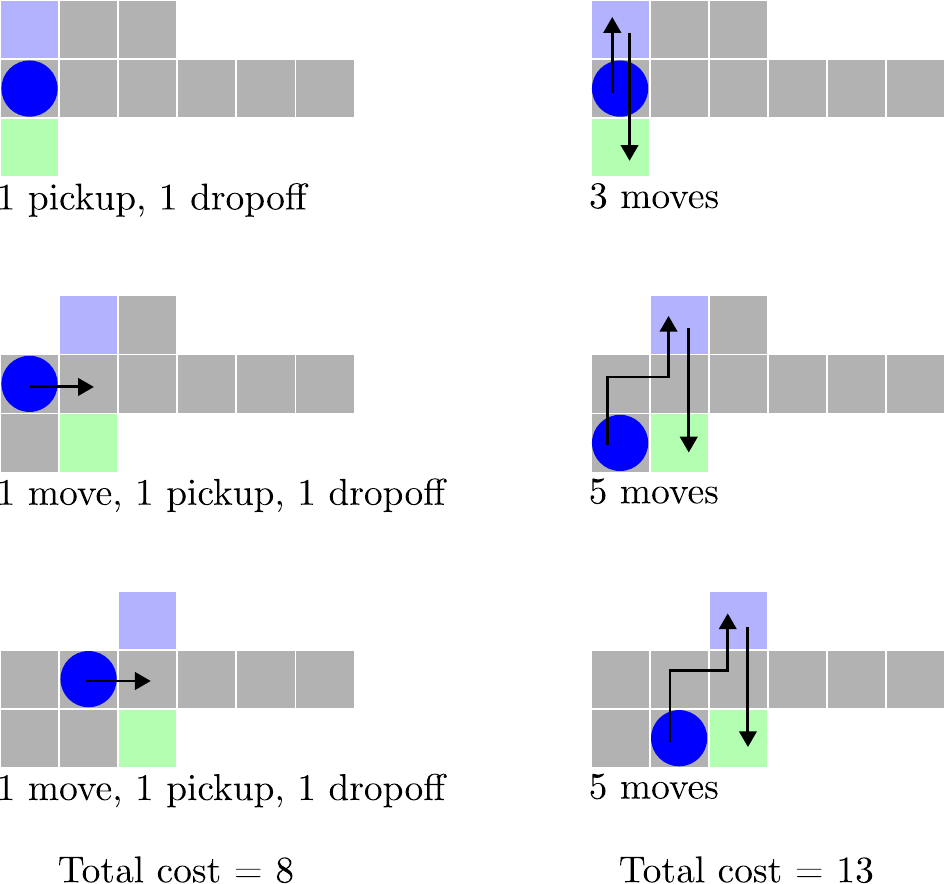}
        \caption{\label{fig:boundExplanation} Visual explanation of why using TSP to move between pickup (blue) and dropoff (green) locations does not provide an accurate lower bound.
        }
    \end{figure}

    We compare the reconfiguration method described above, which we refer to as \CHHC in the remainder of this paper, to the two algorithms presented in~\cite{single-bille-reconfig-IROS} and to an adapted TSP solution.

    The first algorithm we use in our comparison is \mbox{\GLC\!($S,G$)} for \emph{Grow Largest Component}.
    If the start and goal polyominoes $S$ and $G$ do not overlap, the shortest gap between $S$ and $G$ is identified.
    Then, the closest tile to this gap that can be removed without disconnecting $S$ (a~\emph{leaf}) is moved to reduce the gap size by one unit.
    Once there is an overlap, the largest connected component in the overlap is computed.
    All of the positions in $G$ that are reachable from connected component are identified as the set~$N$.
    All of the leaf nodes in $S$ that are not already part of the connected component are identified as the set~$L$.
    Then, a tile from~$L$ closest to an empty position in~$N$ is moved there.
    \GLC is a complete reconfiguration planner.
    One of the key advantages of \GLC is that it minimizes the number of tile pickups and dropoffs.

    The second planner, \mbox{\MWPMexp\!($S,G$)}, uses a minimum-weight perfect matching between all tiles in $S$ and $G$, where distances are computed according to the shortest path using BFS.
    The matching is sorted by distance between the pairs.
    Of all of the leaf nodes in $S$, the one with the longest distance matching is moved as close to its goal destination as possible along the configuration~$S$.
    This planner is not complete and can produce invalid schedules.

    The TSP solution (implemented using the Concorde solver) includes the distance to travel from the~$i$\textsuperscript{th} pickup to the $i$\textsuperscript{th} dropoff and from there, for~${i\in [1,n-1]}$, to the $(i+1)$\textsuperscript{th} pickup.
    For the edge weights, the Manhattan distance between pickups as dropoffs is used.
We add an auxiliary node with directed edges to every tile in~$C_s$ and directed edges from every tile in~$C_t$.
    All of these edges have zero cost.
    Additionally, the TSP cost is modified by subtracting~$2n$ because in a practical application, the robot can pick tiles up and drop them off in adjacent spaces without moving, as demonstrated in \cref{fig:boundExplanation}.
    While the TSP does not require the robot to move on the tiles and disregards connectivity, it is a good baseline to visualize the performance of the three algorithms being tested.
However, it is important to note that the solution is not a strict lower bound.
In particular, it is possible to create instances where the TSP cost is higher than the \CHHC cost, since the latter repeatedly moves tiles in groups which can save movement costs.

    \subsection{Disjoint Configurations}

    \begin{figure*}
        \centering
        \begin{subfigure}{0.5\columnwidth}%
            \includegraphics[page=1,width=\columnwidth]{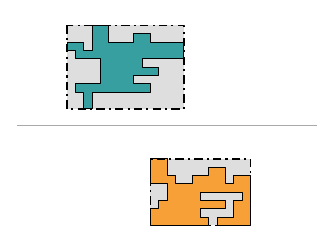}
            \caption{Horizontal bisector}
            \label{fig:bounding-box-overlap-horizontal-bisector}
        \end{subfigure}
        \begin{subfigure}{0.5\columnwidth}%
            \includegraphics[page=2,width=\textwidth]{figures/bounding-box-overlap}
            \caption{Vertical bisector}
            \label{fig:bounding-box-overlap-vertical-bisector}
        \end{subfigure}
        \begin{subfigure}{0.5\columnwidth}%
            \includegraphics[page=3,width=\textwidth]{bounding-box-overlap}
            \caption{Two bisectors}
            \label{fig:bounding-box-overlap-two-bisectors}
        \end{subfigure}
        \begin{subfigure}{0.5\columnwidth}%
            \includegraphics[page=4,width=\textwidth]{bounding-box-overlap}
            \caption{Intersecting bounding boxes}
            \label{fig:bounding-box-overlap-intersecting}
        \end{subfigure}
        \caption{The four possible cases of separation of the start and target configuration.}
        \label{fig:bounding-box-overlap}
    \end{figure*}

    \begin{figure}[!t]
        \centering
        \includegraphics[width=\columnwidth, trim={0cm 6cm 1cm 6cm},clip]{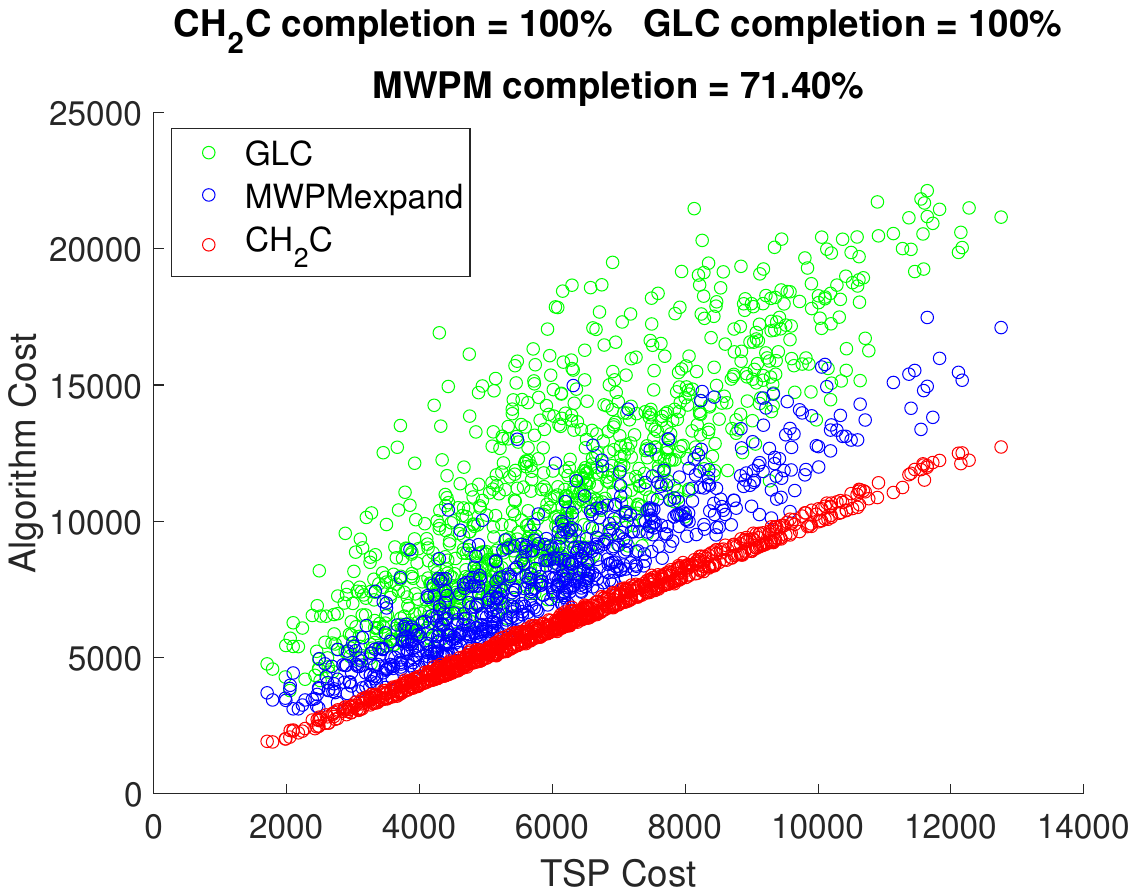}
        \includegraphics[width=.48\columnwidth, trim={2cm 7cm 2cm 7cm},clip]{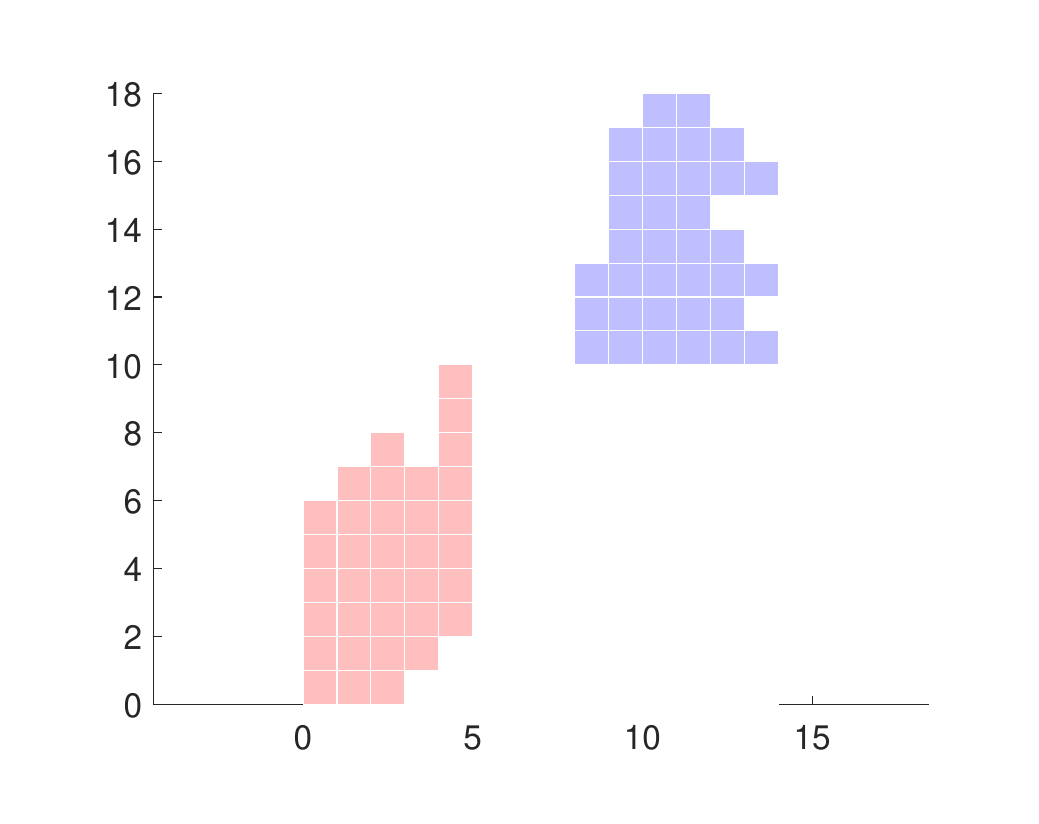}
        \includegraphics[width=.48\columnwidth, trim={2cm 7cm 2cm 7cm},clip]{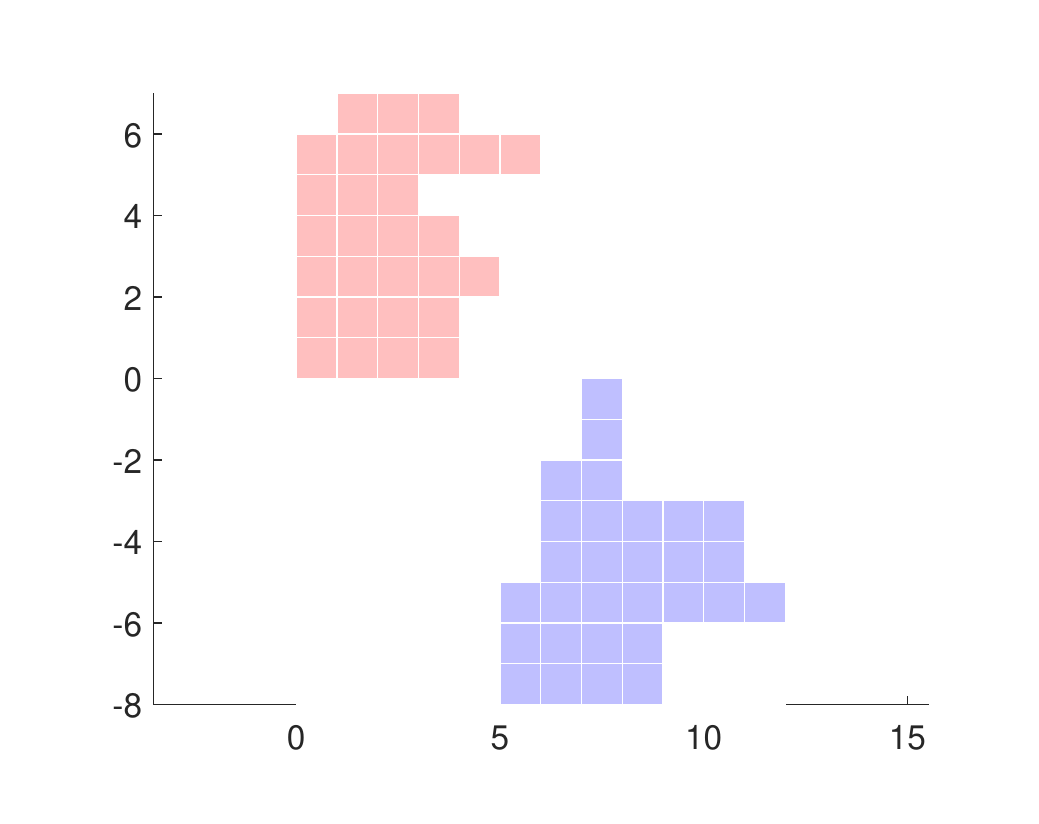}
        \caption{
            \label{fig:sim-round-1}
            Simulation results for test set \boxy.
\CHHC displays a strongly linear relationship with the TSP cost.
            \MWPMexp fails often, and \GLC completes all maps but has the highest costs.
            Example maps are shown.
        }
    \end{figure}

    \begin{figure}[!t]
        \centering
        \includegraphics[width=\columnwidth, trim={0cm 6cm 1cm 6cm},clip]{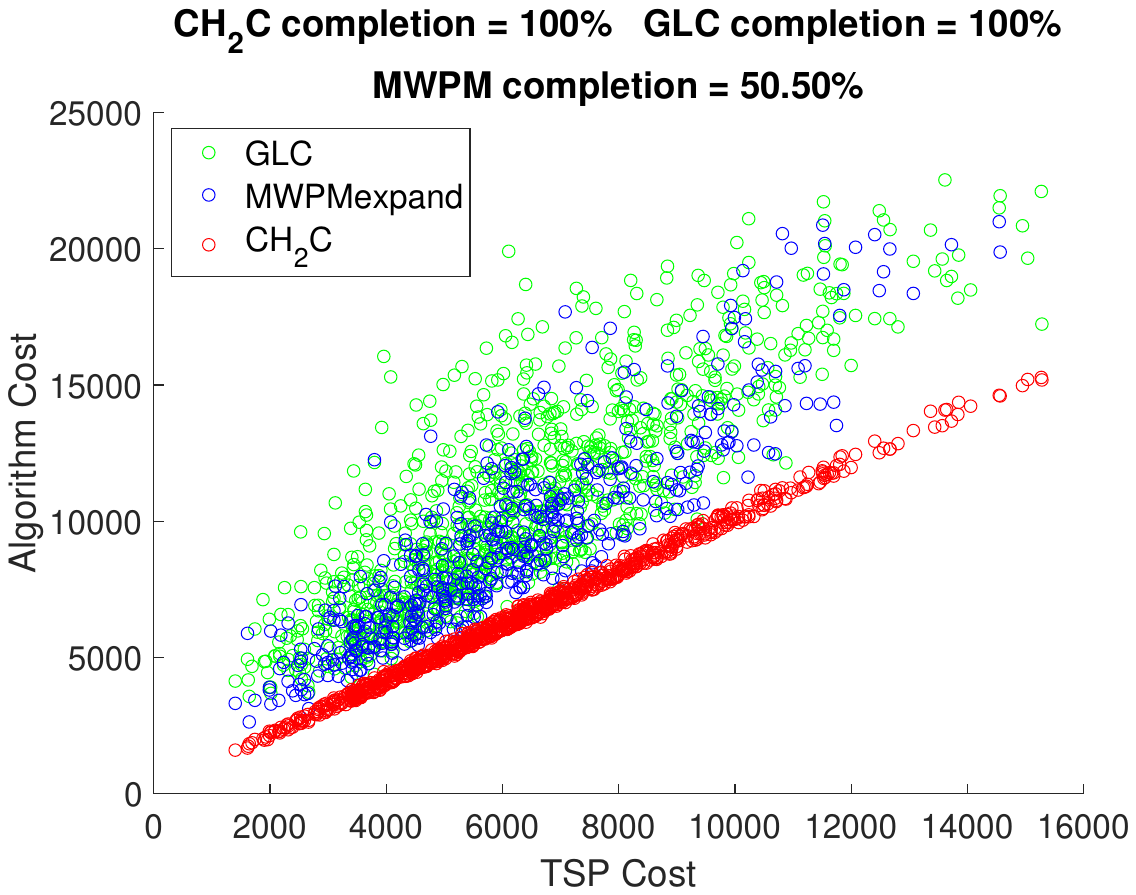}
        \includegraphics[width=.48\columnwidth, trim={2cm 7cm 2cm 7cm},clip]{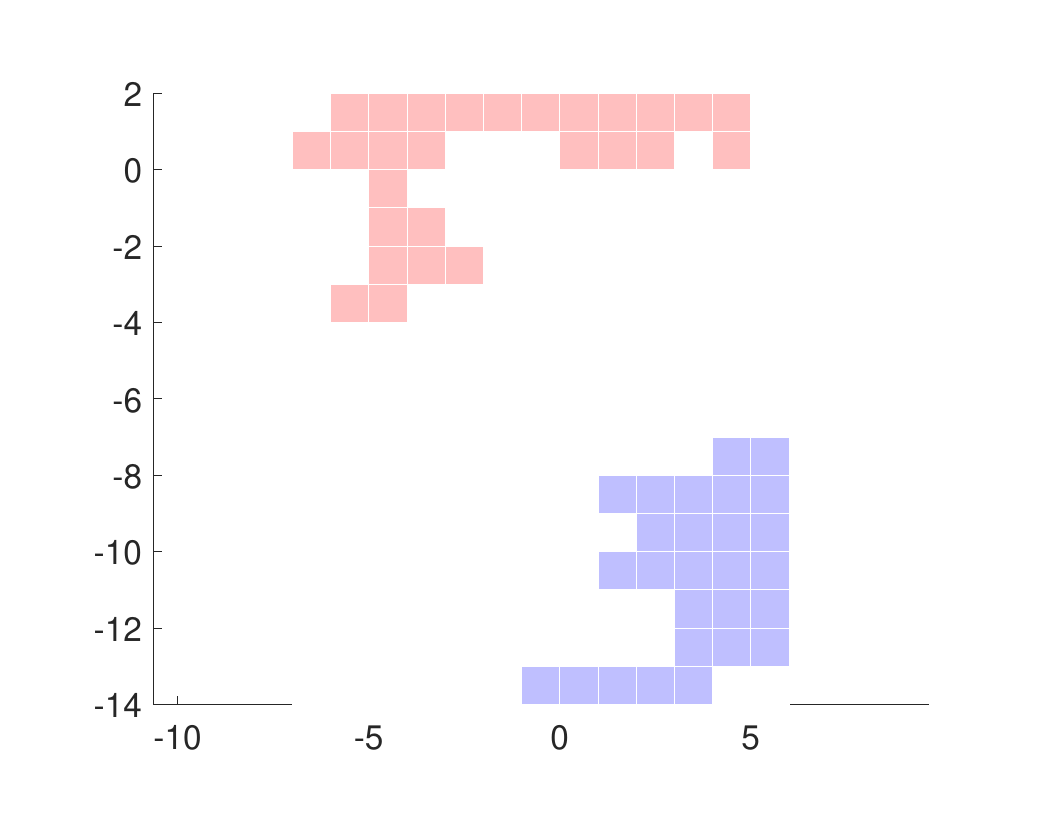}
        \includegraphics[width=.48\columnwidth, trim={2cm 7cm 2cm 7cm},clip]{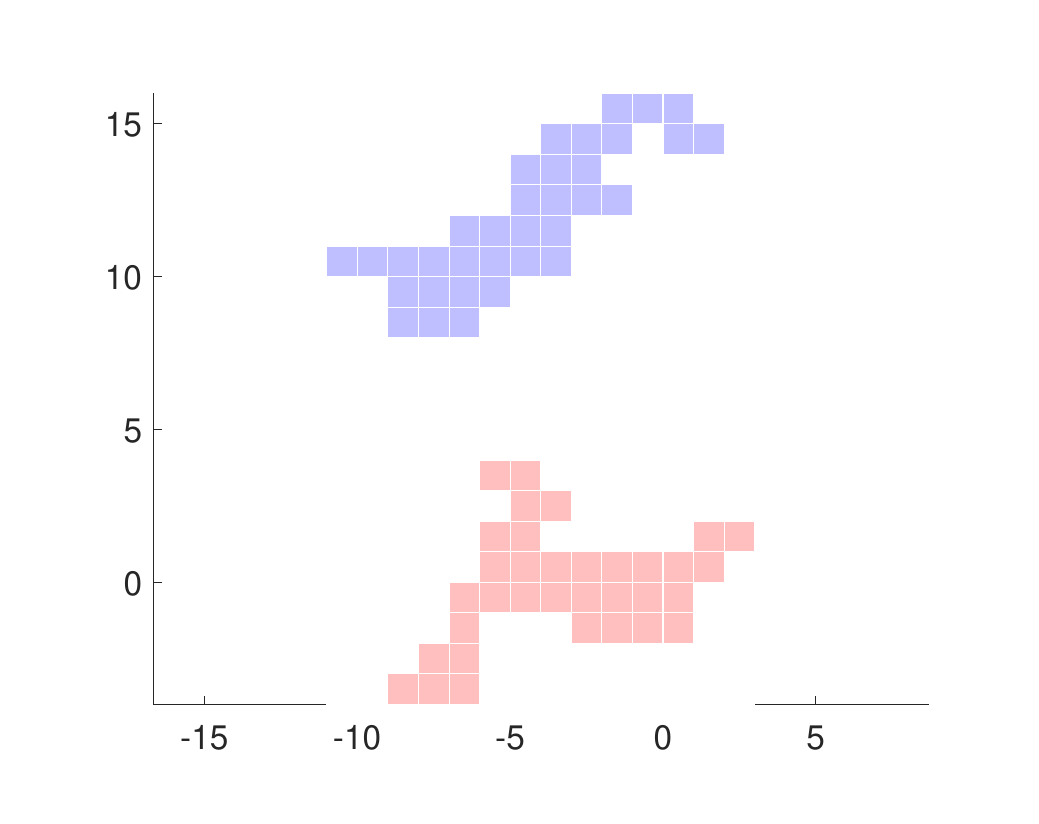}
        \caption{
            \label{fig:sim-round-2}
            Simulation results for test set \snakey, with more complex maps.
            \CHHC performs similarly to the \boxy set.
\MWPMexp can only solve around half of the generated maps and \GLC still tends to return the highest costs.
        }
    \end{figure}

    To evaluate the algorithms, two sets of 1000 maps were created.
All maps belong to the cases shown in \cref{fig:bounding-box-overlap-horizontal-bisector,fig:bounding-box-overlap-two-bisectors}.
For the first set, \boxy, tiles were randomly added to the right and top of existing tiles, resulting in blocky polyominoes.
For the second set, \snakey, tiles were randomly added in all directions next to existing tiles, prioritizing recently placed tiles, resulting in more complex polyominoes.

    \cref{fig:sim-round-1,fig:sim-round-2} show the total cost of the first two simulation rounds for the three algorithms on the $y$-axis.
This cost includes the number of moves with and without a load, as well as the total number of pickups and dropoffs.
All of these operations are considered to have unit cost.
The TSP costs are used for the $x$-axis.
Example maps are also shown below.

    \subsection{Overlapping Configurations}

    \begin{figure}[t]
        \centering
        \includegraphics[width=\columnwidth, trim={0cm 6cm 1cm 6cm},clip]{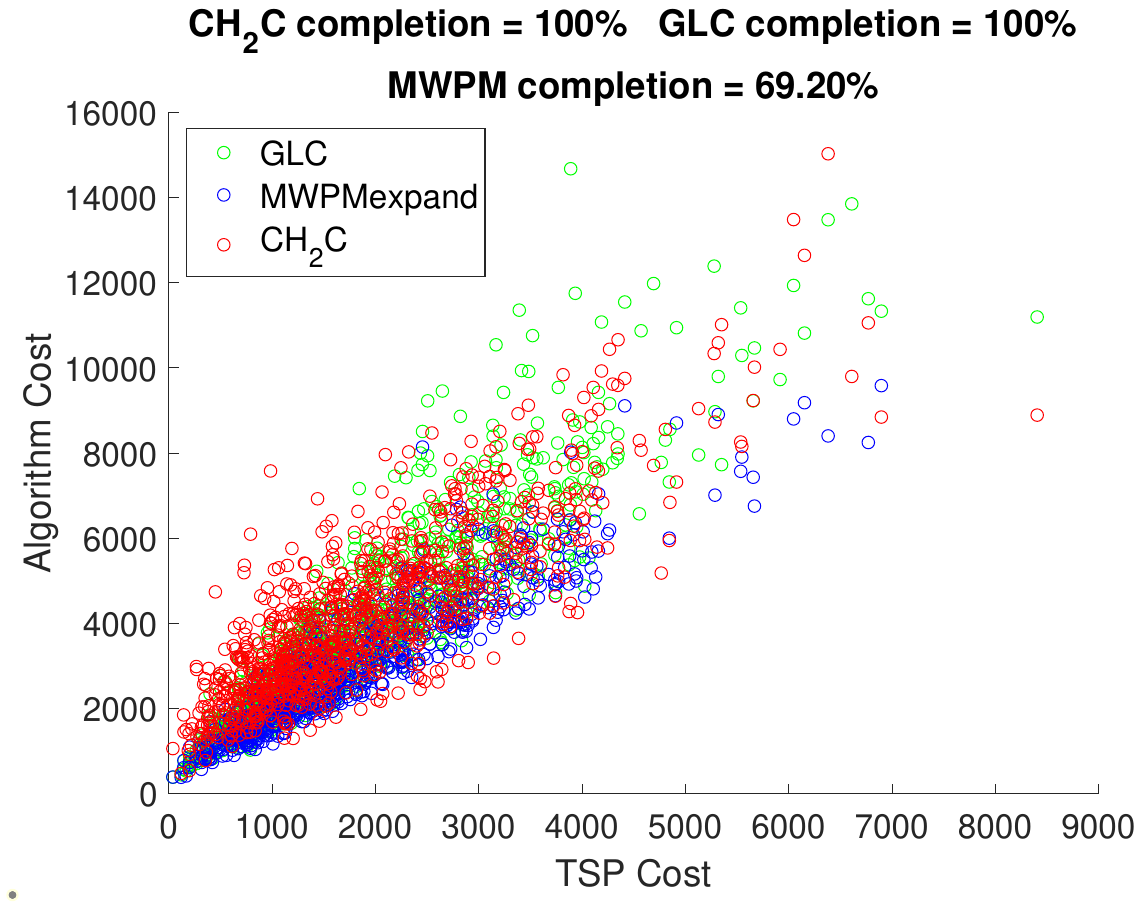}
        \includegraphics[width=.48\columnwidth, trim={2cm 7cm 2cm 7cm},clip]{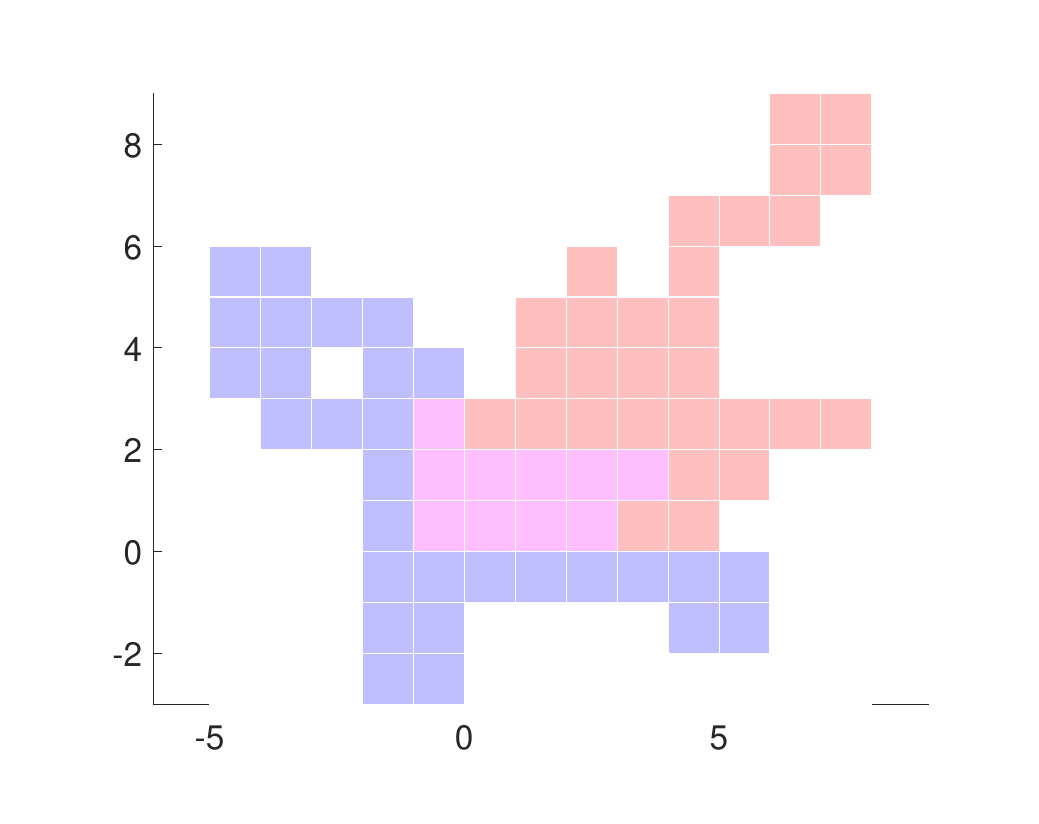}
        \includegraphics[width=.48\columnwidth, trim={2cm 7cm 2cm 7cm},clip]{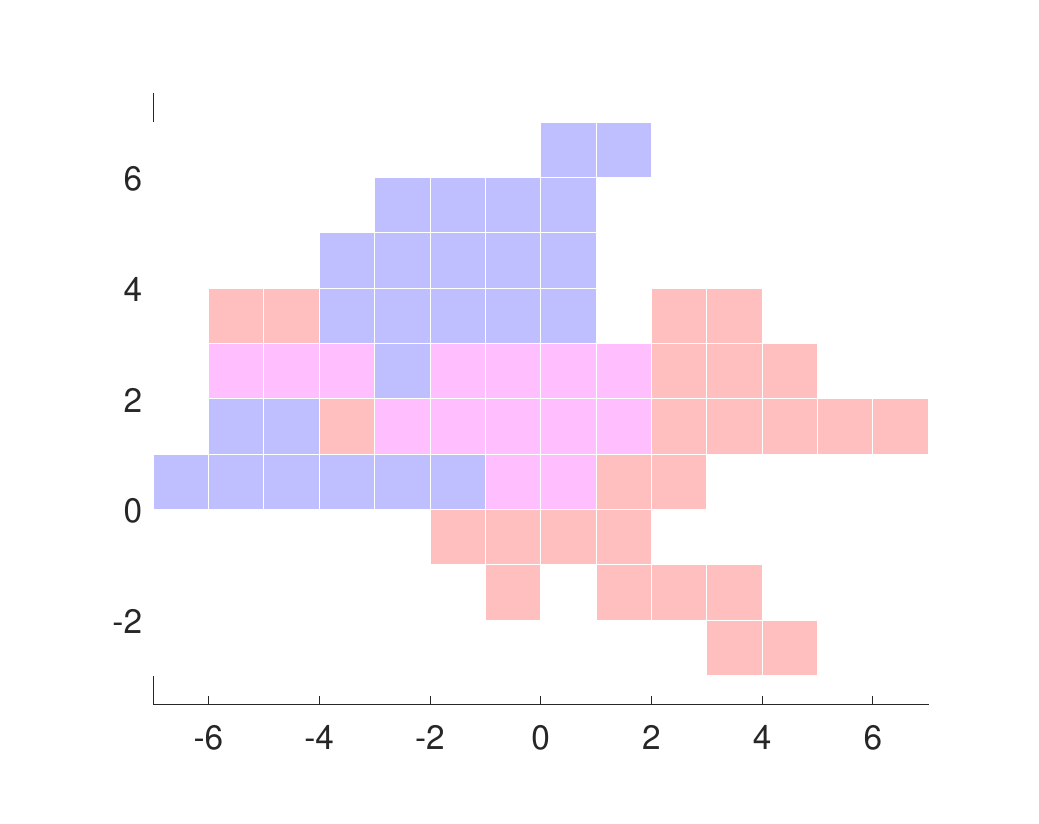}
        \caption{
            \label{fig:sim-round-3}
            Simulation results for test set \overlapping.
This time, all the maps are created to have overlap.
As a result, \CHHC no longer performs much better than \GLC, but at least completes all maps unlike \MWPMexp.
        }
    \end{figure}

    \CHHC performs well when the bounding boxes of the two configurations do not overlap.
    In this case, the intermediate histograms face opposite directions and tiles can be moved on shortest paths to their target locations.
    While the algorithm can be easily adapted for overlapping configurations, tiles are no longer necessarily moved on optimal paths.
    To demonstrate, \overlapping, a third set of 1000 maps with overlapping polyominoes (\cref{fig:bounding-box-overlap-intersecting}) was created.
    The results are depicted in \cref{fig:sim-round-3}.
    The performance drops to the level of \GLC.

    \subsection{Number of Pickups and Dropoffs}

    \begin{figure}[t]
        \centering
        \includegraphics[width=\columnwidth, trim={1cm 7cm 1cm 7cm},clip]{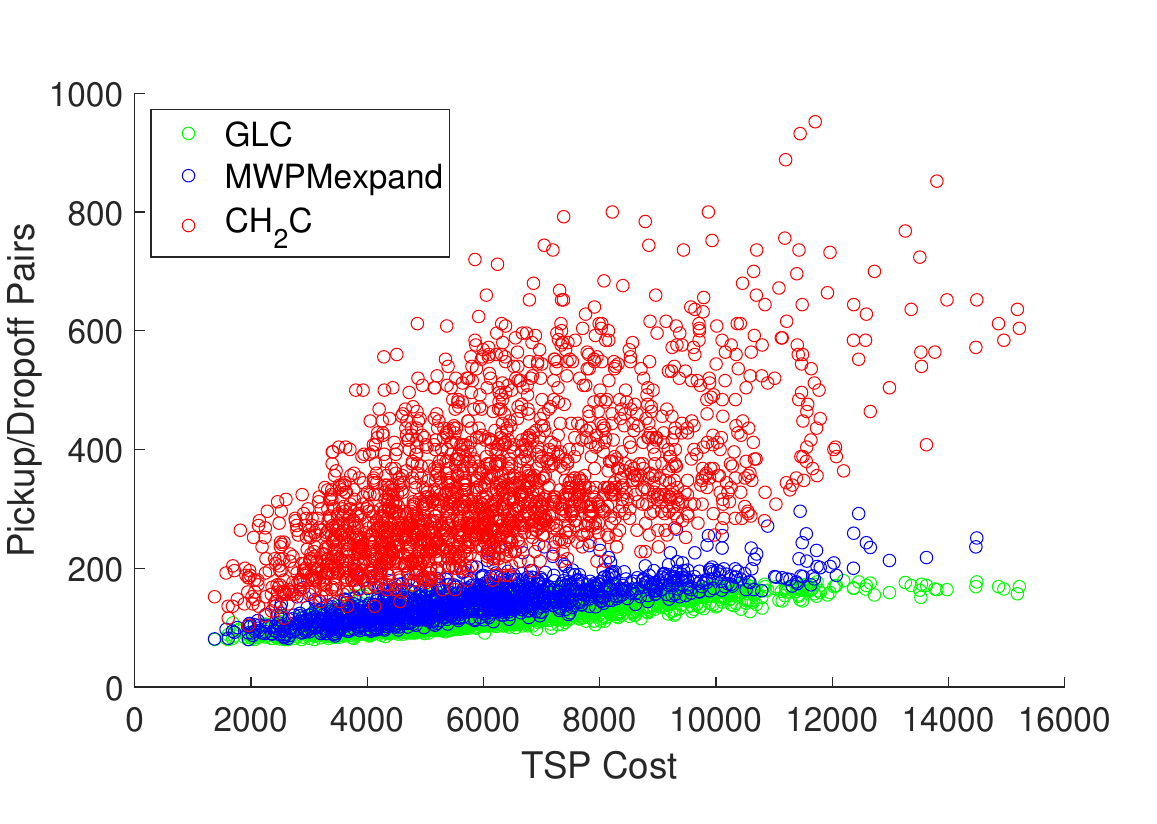}
        \caption{
            \label{fig:pu-do-pairs}
            Number of pickup/dropoff pairs for the different algorithms as a function of TSP cost, for all maps in \boxy and \snakey.
If pickup/dropoff cost is high compared to moving cost, \CHHC performance worsens.
        }
    \end{figure}

    Although \CHHC significantly outperforms \MWPMexp and \GLC in the previous tests, this happens only when the costs for moving the robot and picking up/dropping off tiles are equal.
As seen in \cref{fig:pu-do-pairs} (\boxy and \snakey sets combined), \CHHC results in significantly more tile pickups and dropoffs due to the histogram creation process.
    As expected, \GLC performs best since it minimizes the number of tile pickups and dropoffs by design, with \MWPMexp trailing close behind.
    Thus, if picking up or placing tiles has significantly higher cost than moving, \MWPMexp or \GLC yield more efficient schedules than \CHHC.

    \subsection{Free Component Strategy}

    \begin{figure}[t]
        \centering
        \includegraphics[width=.48\columnwidth, trim={3cm 8cm 3cm 8cm},clip]{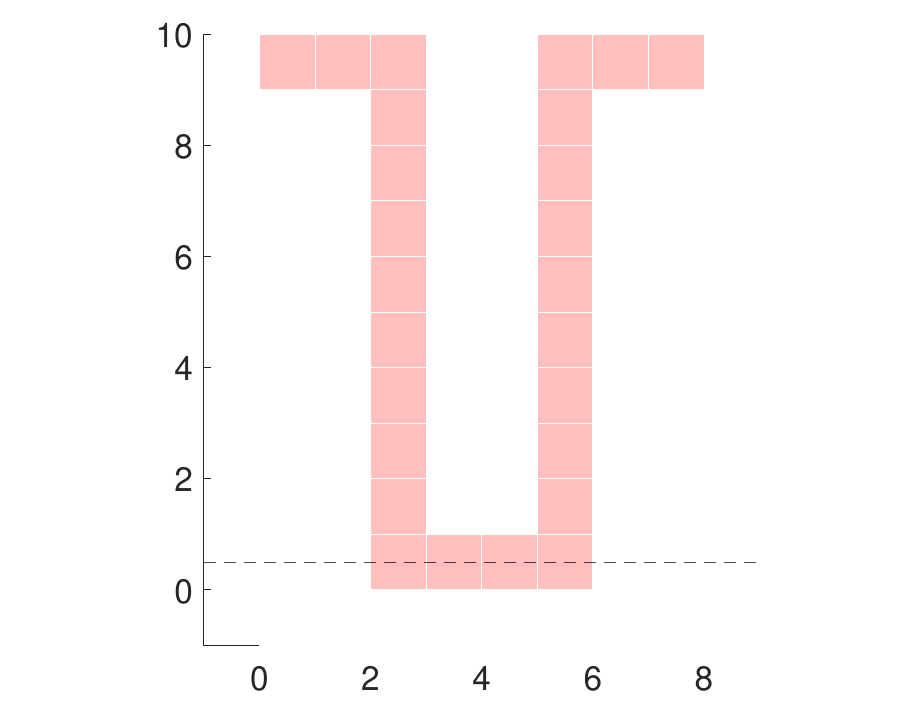}
        \includegraphics[width=.48\columnwidth, trim={3cm 8cm 3cm 8cm},clip]{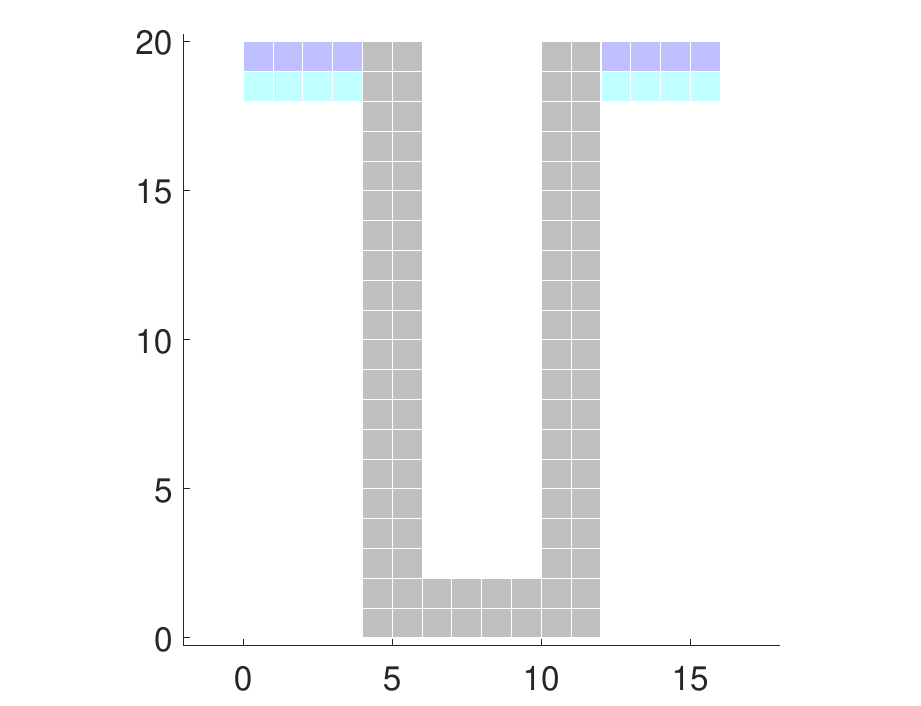}
        \includegraphics[width=.48\columnwidth, trim={3cm 8cm 3cm 8cm},clip]{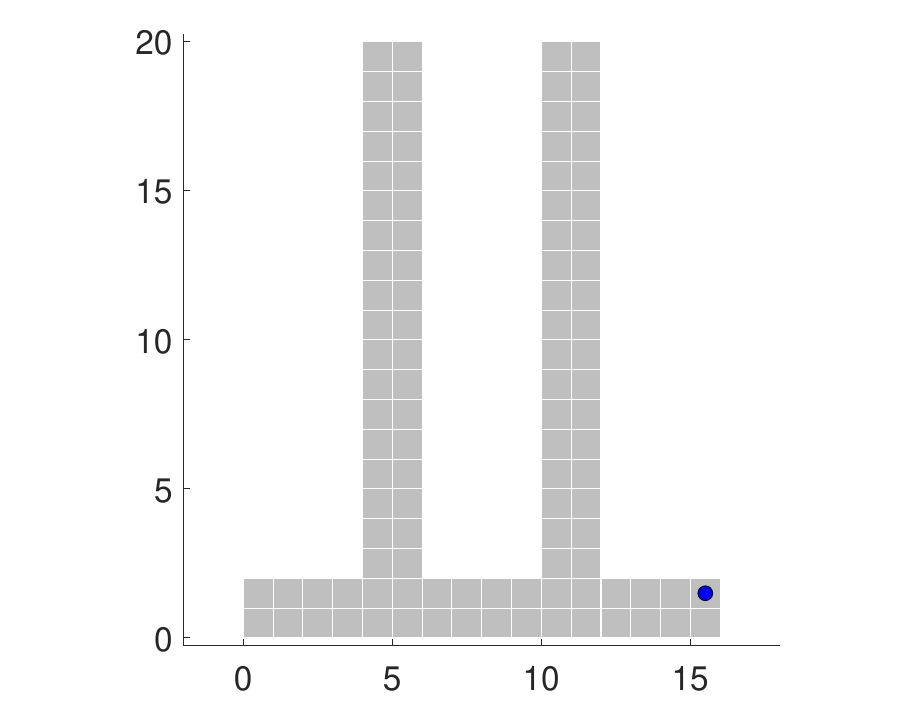}
        \includegraphics[width=.48\columnwidth, trim={1cm 0cm 1cm 1cm},clip]{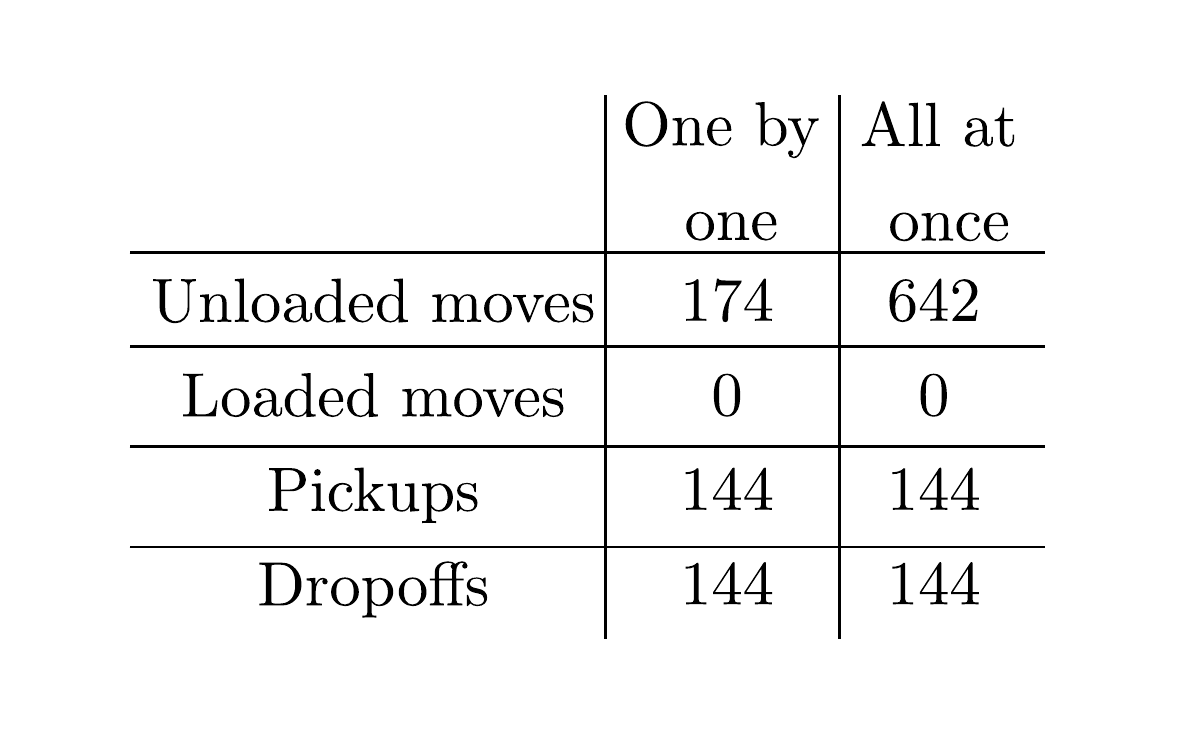}
        \caption{
            \label{fig:u-shape}
            (Top-left) Map with U-shaped starting configuration.
(Top-right) Free components, left and right, highlighted in blue.
(Bottom-left) Histogram created from starting configuration.
(Bottom-right) Costs of histogram creation using different strategies for moving the free components.
        }
    \end{figure}

    Next, we compare the effect of moving free components one by one vs.\ all at once (the previous simulations use the former strategy as designed by Becker et al.~\cite{becker2025moving}).
An example of a map with a large cost difference is shown in \cref{fig:u-shape}.
    If the free components are moved together, the robot must get from the left side to the right side multiple times, racking up the unloaded move costs.
Note that loaded move, pickup and dropoff costs are unaffected.

    \subsection{Horizontal vs.\ Vertical Bisector}

    \begin{figure}[t]
        \centering
        \includegraphics[width=\columnwidth, trim={0.5cm 7cm 1cm 7cm},clip]{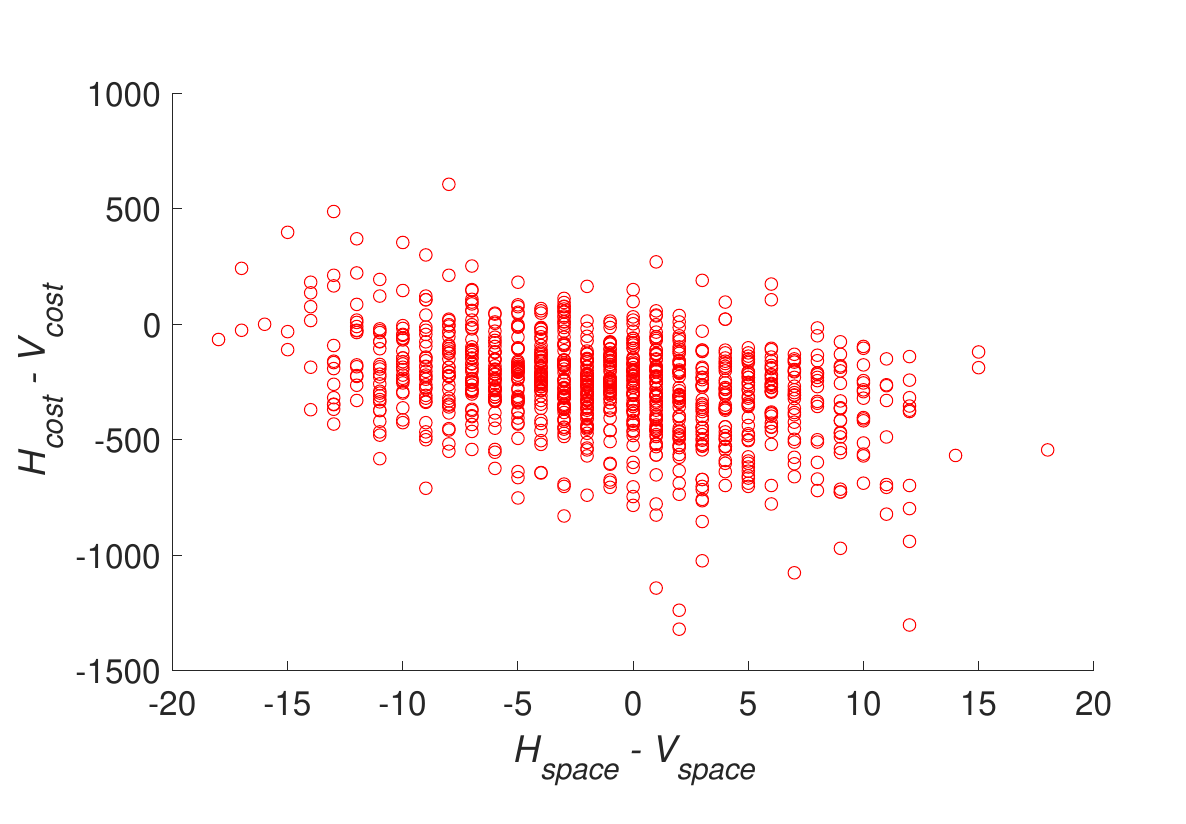}
        \caption{
            \label{fig:bisectorTest}
            Cost difference when using \CHHC with a horizontal and a vertical bisector.
The $x$-axis shows the difference between the horizontal and vertical gaps between the starting and target configurations' bounding boxes.
        }
    \end{figure}

    Finally, we consider the impact of the histogram orientation (north/south or east/west).
    This was done by creating a new set of 1000 maps, in which the starting and target configurations do not share $x$ or $y$ coordinates (\cref{fig:bounding-box-overlap-two-bisectors}).
\CHHC is run twice, first regularly and then with the maps rotated by 90 degrees, which has the same effect as using a vertical bisector line and creating east/west histograms.
    The results are shown in \cref{fig:bisectorTest} as the difference of the horizontal and vertical bisector costs, $H_{\mathit{cost}}$ and $V_{\mathit{cost}}$, relative to the difference between the horizontal and vertical separations between the configurations' bounding boxes, $H_{\mathit{space}}$ and $V_{\mathit{space}}$.
    Although the trend is not strong, the longest separation between the bounding boxes can be used to dictate which bisector line to use (horizontal for configurations that are close vertically and vice versa).

    \section{Hardware Demonstration} \label{sec:Results}

    We utilized the \BILLE bot platform~\cite{jenett2019material} to implement the \CHHC algorithm.
    This \mbox{6-DoF} inchworm-type robot can only walk on tiles, so the connectivity requirement must be enforced so that all the tiles can be reached.
    Using a gripper on the front foot, the robot can move tiles one at a time.

\begin{figure}[t]
        \centering
        \definecolor{myRed}{RGB}{252,130,130}
        \definecolor{myYellow}{RGB}{253,226,134} \definecolor{myGreen}{RGB}{132,252,133}
        \definecolor{myBlue}{RGB}{129,131,251}
        \definecolor{myLavender}{RGB}{225,133,252}
        \definecolor{myCyan}{RGB}{133,226,253}
        \adjincludegraphics[width=0.8\columnwidth] {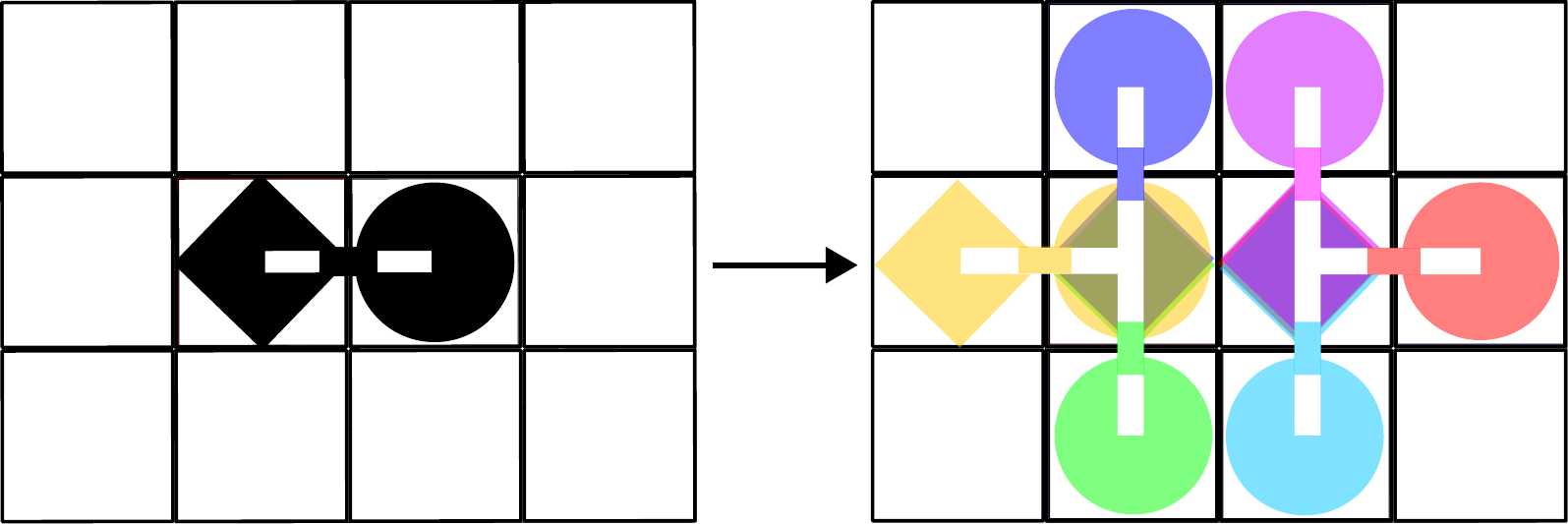}
        \caption{\label{fig:Neighbors}
        Starting from a configuration (left), our \BILLE bot can reach several configurations after a single move (right).
These moves are
        \textcolor{myRed}{$\bullet$} stepping forward,     \textcolor{myYellow}{$\bullet$} backward,     \textcolor{myGreen}{$\bullet$} moving the front foot 90\textdegree~cw or \textcolor{myBlue}{$\bullet$} ccw, and     \textcolor{myLavender}{$\bullet$} stepping forward and moving the front foot 90\textdegree~cw or    \textcolor{myCyan}{$\bullet$} ccw.
        The robot can also turn 180\textdegree~cw or ccw (not shown for legibility).}
    \end{figure}

\begin{figure}[t]
        \centering
        \includegraphics[width=\columnwidth, trim={0 0 0 0},clip]{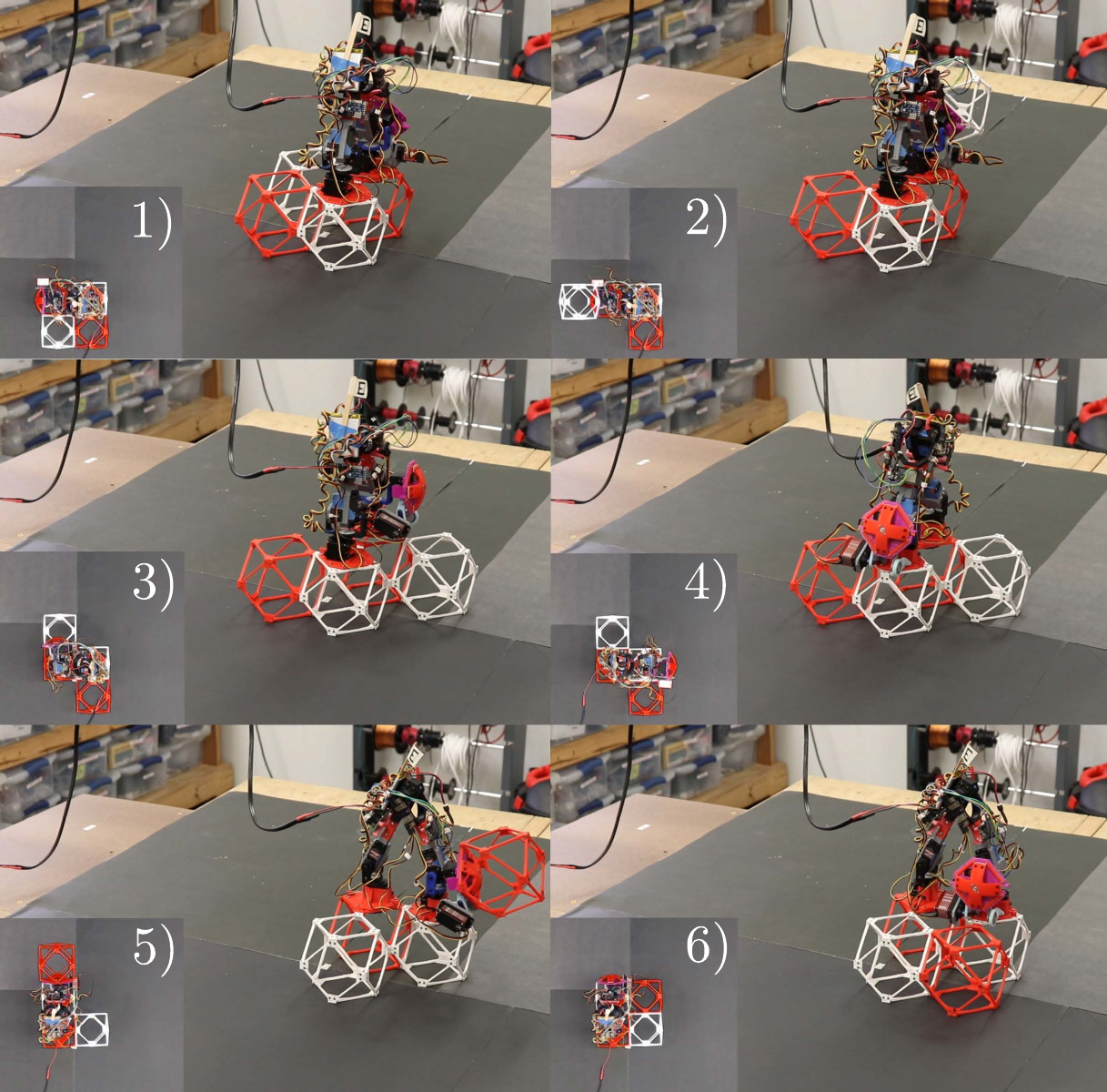}
        \caption{\label{fig:RaminSimple} The \BILLE bot translating a $2 \times 2$ configuration sideways.
The bottom-left of each frame shows a top view of the map and robot.
        }
    \end{figure}

    There are some key differences between the \BILLE bot and the idealized tile mover used in the theory and simulation sections.
The \BILLE bot has two feet and an orientation.
    Both feet must be connected to the structure before the robot can remove or place a tile.
    These factors generate constraints on foot placement and move cycles when moving to pickup and dropoff locations.
    However, the robot is still very flexible in terms of available moves (see \cref{fig:Neighbors}), and can translate a polyomino with limited space, as shown in \cref{fig:RaminSimple}.

    Another implementation challenge lies with the validity of tiles to be moved at a given time.
    The algorithm does not discriminate as long as connectivity is kept, which often results in pickup and dropoff locations connected to more than one tile.
    In practice, the \BILLE bot struggles with pickup and dropoff locations surrounded by two or more tiles, due to the strength of the magnets holding the tiles together being too strong.
    A potential solution is using fasteners instead of magnets to hold the voxels, such as in~\cite{2023-MMIC-OliviaFormoso}.
    The \BILLE also cannot pick or place a tile if there is another tile in front (with respect to the orientation of the robot) of the desired location.
    Therefore, for the demo in \cref{fig:conceptImg}, the robot was manually assisted with lifting and placing some tiles.

    \section{Conclusions}\label{sec:conclusions}

    In this paper we tested the performance of a new reconfiguration algorithm (\CHHC) in simulation, comparing it to two previous algorithms (\GLC and \MWPMexp).
\CHHC results in lower costs when the starting and target configurations can be separated by a horizontal or vertical line, and pickup/dropoff costs are equal to moving costs.
    However, since it performs many more tile pickups and dropoffs than the other two algorithms, increasing the cost of these operations makes it less appealing.
    The algorithm was also demonstrated in hardware using an inchworm-type robot (with assistance for picking up and placing tiles).

    There are many avenues for future work.
    While \CHHC computes schedules within a constant factor of an optimal solution for separated configurations, this does not hold if the configurations overlap.
    Additionally, this method does not handle obstacles, unlike \GLC and \MWPMexp.
    It is unclear how the 2D histogram method could be adapted to 3D, as many tiles may be unreachable from the outside of the shape and auxiliary tiles may be necessary to maintain connectivity~\cite{miltzow2020hiding}.
    \CHHC could be augmented to adhere to real-world tile placement constraints specific to different robotic platforms.
    Finally, we are exploring both the multi-robot variant and new algorithmic ideas to circumvent hardware limitations.

    \bibliographystyle{IEEEtranDOI}
    \bibliography{biblio}
\end{document}